%% file: acl2021.tex
\documentclass[11pt,a4paper]{article}

\usepackage{authblk}

\usepackage[hyperref]{acl2021}
\usepackage{times}
\usepackage{latexsym}

\usepackage{microtype}

\usepackage{inconsolata}
\usepackage{graphicx}
\usepackage{caption}
\usepackage{subcaption}
\usepackage{multirow}
\usepackage{hhline}
\usepackage{amsmath}
\usepackage{amsfonts}
\usepackage{xcolor}
\usepackage{soul}
\usepackage{algorithm}
\usepackage{algpseudocode}
\usepackage[normalem]{ulem}
\usepackage{pifont}
\newcommand{\cmark}{\ding{51}}%
\newcommand{\xmark}{\ding{55}}%
\usepackage{xcolor}
\aclfinalcopy % Uncomment this line for the final submission
\title{Complex Reading Comprehension Through Question Decomposition}

\author[ ]{Xiao-Yu Guo}
\author[ ]{Yuan-Fang Li}
\author[ ]{Gholamreza Haffari}
\affil[ ]{Faculty of Information Technology, Monash University, Melbourne, Australia}
\affil[ ]{\url{{xiaoyu.guo, yuanfang.li, gholamreza.haffari}@monash.edu}}

\begin{document}
\maketitle
\begin{abstract}
Multi-hop reading comprehension requires not only the ability to reason over raw text but also the ability to combine multiple evidence. 
We propose a novel learning approach that helps language models better understand difficult multi-hop questions and perform ``complex, compositional'' reasoning.
Our model first learns to decompose each multi-hop question into several sub-questions by a trainable \emph{question decomposer}.
Instead of answering these sub-questions, we directly concatenate them with the original question and context, and leverage a \emph{reading comprehension} model to predict the answer in a sequence-to-sequence manner.
By using the same language model for these two components, our best \emph{seperate}/\emph{unified} t5-base variants outperform the baseline by 7.2/6.1 absolute F1 points on a hard subset of DROP dataset.
\end{abstract}

\input{sec1-intro}

\input{sec2-relat}

\input{sec3-model}

\input{sec4-exper}

\input{sec5-concl}

% \input{sec6-limit}

%\section*{Acknowledgements}

% Entries for the entire Anthology, followed by custom entries
\bibliography{acl2021}
\bibliographystyle{acl_natbib}

\end{document}

%% file: sec1-intro.tex
%!TEX ROOT = ./acl2021.tex
\section{Introduction}
\label{sec1-intro}

Multi-hop Reading Comprehension (RC) is a challenging problem that requires compositional, symbolic and arithmetic reasoning capabilities. 
Facing a difficult question, humans tend to first decompose it into several sub-questions whose answers can be more easily identified. 
The final answer to the overall question can then be concluded from the aggregation of all sub-questions' answers. 
For instance, for the question in Table \ref{tab:example}, we can naturally decompose it into three simpler sub-questions (1) ``return the touchdown yards'', (2) ``return the fewest of $\#1$'', and (3) ``return who caught $\#2$''. 
The tokens $\#1$ and $\#2$ are the answers to the first and second sub-questions respectively. Finally, the player with the touchdown of $\#2$ is returned as the final answer. 

\begin{table}[!htb]
    \centering
    \begin{tabular}{ll}
         \hline
         $\mathbf{C}$
         & First, Detroit's Calvin Johnson caught \\
         & a 1-yard pass in the third quarter. The \\
         & game's final points came when Mike \\
         & Williams of Tampa Bay caught a 5-yard.  \\ \hline
         $\mathbf{Q}$ & Who caught the touchdown for the \\
         & fewest yards? \\ \hline
         $\mathbf{Q_1}$ & return the touchdown yards \\
         $\mathbf{Q_2}$ & return the fewest of $\#1$ \\ 
         $\mathbf{Q_3}$ & return who caught $\#2$ \\ \hline
         $\mathbf{A}$ & Calvin Johnson \\ \hline
    \end{tabular}
    \caption{An example for reading comprehension. $\mathbf{C}$ is the context, $\mathbf{Q}$ is a hard multi-hop question, and $\mathbf{Q_1}$, $\mathbf{Q_2}$, $\mathbf{Q_3}$ are sub-questions annotated in \textsc{Break} dataset. $\mathbf{A}$ is the answer to $\mathbf{Q}$.}
    \label{tab:example}
\end{table}

State-of-the-art RC techniques employ large-scale pre-trained language models (LMs) such as GPT-3~\cite{lms-gpt-3} for their superior representation and reasoning capablities. 
Chain of thought prompting~\cite{chain-of-thought} elicits strong reasoning capability of LMs by providing intermediate reasoning steps. 
Least-to-most prompting~\cite{least-to-most} further shows the feasibility of conducting decomposition and multi-hop reasoning, which happen on the decoder side together with the answer prediction procedure.
However, compared to supervised learning models, both of these methods rely on extremely large LMs with tens and hundreds of \textbf{billions} of parameters to achieve competitive performance, thus requiring expensive hardware and incurring a large computation footprint.

Despite significant research on RC \cite{DROP:2019,Unsupervised:2020}, those questions that require strong compositional generalisability and numerical reasoning abilities are still challenging to even the state-of-the-art models \cite{ran-etal-2019-numnet,chen-etal-2020-qdgat,ChenLYZSL20-nerd,chain-of-thought,least-to-most}. 
While decomposition is a natural approach to tackle this problem, the lack of sufficient ground-truth sub-questions limits our ability to train RC models based on large LMs. 

In this paper, we propose a novel low-budget (only  1\textperthousand~parameters of GPT-3) learning approach to improve LMs' performance on hard multi-hop RC such as the Break subset of DROP \cite{DROP:2019}. 
Our model consists of two main modules: (1) an encoder-decoder LM as a \emph{question decomposer} and (2) another encoder-decoder LM as the \emph{reading comprehension} model. 
First, we train the question decomposer to decompose a difficult multi-hop question to sub-questions from a limited amount of annotated data.
Next, instead of solving these sub-questions, we train the reading comprehension model to predict the final answer by directly concatenating the sub-questions with the original question. 
We further propose a \emph{unified} model that utilizes the same LM for both question decomposition and reading comprehension with task-specific prompts.
With 9$\times$ weakly supervised data, we design a Hard EM-style algorithm to iteratively optimise the \emph{unified} model.

To prove the effectiveness of our approach, we leverage two different types of LMs: T5~\cite{lms-t5} and Bart~\cite{lms-bart} to build baselines and our variants.
The experimental results show that without changing the model structure, our proposed variant outperforms the end-to-end baseline.
By adding ground-truth sub-questions, gains on the F1 metric are 1.7 and 0.7 using T5 and Bart separately. 
Introducing weakly supervised training data can help improve the performance of both \emph{separate} and \emph{unified} variants by at least 4.4 point on F1.
And our method beats the state-of-the-art model GPT-3 by a large margin.

%% file: sec2-relat.tex
%!TEX ROOT = ./acl2021.tex
\section{Related Work}
\label{sec2-relat}

\paragraph{Multi-hop Reading Comprehension} mentioned in this paper requires more than one reasoning or inference step to answer a question.
For example, multi-hop RC in DROP \cite{DROP:2019} requires numerical reasoning such as addition, subtraction.
To address this problem, \citeauthor{DROP:2019} proposed a number-aware model NAQANet that can deal with such questions for which the answer cannot be directly extracted.
NumNet~\cite{ran-etal-2019-numnet} leveraged Graph Neural Network to design a number-aware deep learning model.
QDGAT~\cite{chen-etal-2020-qdgat} distinguished number types more precisely by adding the connection with entities and obtained better performance.
Nerd~\cite{ChenLYZSL20-nerd} searched possible programs exhaustively based on the ground-truth and employed these programs as weak supervision to train the whole model.

\paragraph{Question Decomposition} 
is the approach that given a complex question, break it into several simple sub-questions.
These sub-questions can also be Question Decomposition Meaning Representation (QDMR)~\cite{break-it-down} for complex questions.
Many researchers~\cite{Unsupervised:2020,decomposition-break-perturb-build} have been trying to solve the problem by incorporating decomposition procedures.
For example, \citet{Unsupervised:2020} propose a model that can break hard questions into easier sub-questions. 
Then, simple QA systems provide answers of these sub-questions for downstream complex QA systems to produce the final answer corresponding to the original complex question. 
\citet{fu-etal-2021} propose a three-stage framework called Relation Extractor Reader and Comparator (RERC), based on complex question decomposition.
Different from these approaches, we aim to improve the multi-hop capability of current encoder-decoder models without dedicated pre-designing the architecture.

\paragraph{Language Models} like BERT~\cite{BERT:2019}, GPT families~\cite{lms-gpt,lms-gpt-2,lms-gpt-3}, BART~\cite{lms-bart} and T5~\cite{lms-t5} are demonstrated to be effective on many NLP tasks, base on either fine-tuning or few-shot learning~\cite{chain-of-thought,least-to-most}, even zero-shot learning. 
However, LMs suffer a lot from solving multi-hop questions and logic reasoning and numerical reasoning problems. 
Although some research~\cite{scratchpads,chain-of-thought} has conducted experiments on either simple or synthetic datasets and shown the effectiveness, \citet{impact-2022} indicates that the model reasoning is not robust enough.

Recently, \citet{lmc-2022} points out that prompted models can be regarded as employing a unified framework a \emph{language model cascade}.
From the perspective view of probabilistic programming, several recent literature~\cite{chain-of-thought,least-to-most} are formalized.
In this paper, we also treat our whole process as a probabilistic model that is consistent to \citet{lmc-2022}. 

%% file: sec3-model.tex
%!TEX ROOT = ./acl2021.tex
% \begin{figure*}[t]
%     \centering
%     \begin{subfigure}[b]{0.49\textwidth}
%         \centering
%         \includegraphics[width=\textwidth]{Figures/Picture 1.png}
%         \caption{Separate Variant: different LMs for Question Decomposer and RC Component}
%         \label{fig:decomposer-rc}
%     \end{subfigure}
%     \begin{subfigure}[b]{0.49\textwidth}
%         \centering
%         \includegraphics[width=\textwidth]{Figures/Picture 2.png}
%         \caption{Unified Variant: Single LM for Question Decomposer and RC Component}
%         \label{fig:t5-unified}
%     \end{subfigure}
%     \caption{The variants of the proposed model.}
%     \label{fig:encoder}
% \end{figure*}

\begin{figure*}[t]
    \centering
    \includegraphics[width=\textwidth]{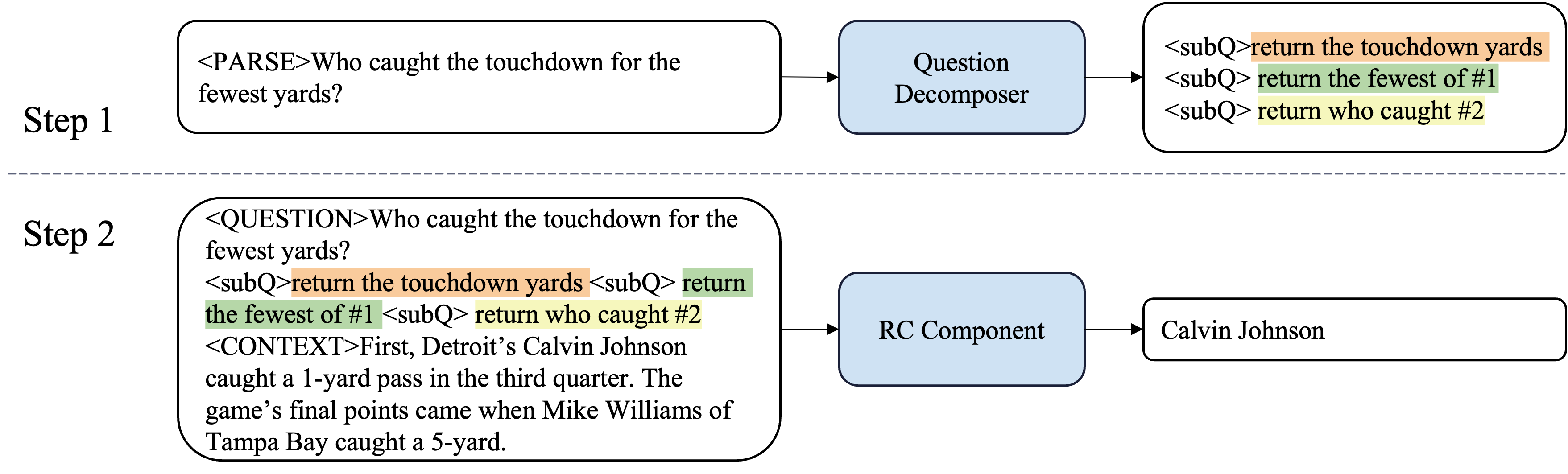}
    \caption{Our model structure on complex reading comprehension through question decomposition. Step 1: Question Decomposer generates a sequence of sub-questions; Step 2: RC component predicts the answer based on question, sub-questions and the given context. The context of this given example is truncated.}
    \label{fig:decomposer-rc}
\end{figure*}

\section{Complex Question Answering Through Decomposition}
\label{sec3-model}
Our focus in this work is on complex questions requiring multi-hop reasoning. 
As such, our approach consists of the following two steps: 
\begin{enumerate}
\item The complex question is decomposed to a sequence of sub-questions. The decomposition of the question is performed by the \emph{question decomposer} component of our system. %, detailed in \S \ref{sec3-prob-model}. 

%First, a question decomposer learns to decompose each difficult multi-hop question into several sub-questions in an end-to-end trainable way. Section \ref{sec3-sub-semantic-parser} details our decomposing approach.
\item The model produces the answer to the complex question leveraging the generated subquestions to provide guidance to the reasoning of the system. This is performed by the \emph{reading comprehension} component. 

%Our model leverages another end-to-end approach that also regards the sub-questions obtained from previous step as the input. Section \ref{sec3-sub-encoder} details our reasoning approach.
\end{enumerate}

We use LMs such as T5 and Bart as the backbone\footnote{Our approach is general, and it can be used with other pre-trained seq2seq models and language models as well.} for {both}  question decomposer and the reading comprehension (Figure \ref{fig:decomposer-rc}). 
We present several variants of our model, depending whether the models for the above two steps are either separate or unified using multitask learning. 
%
%We show that the same multitask-trained T5 can successfully decompose complex questions and make use of the decomposed questions to guide its reasoning process when generating the answer. 
%
As we have the ground truth question decomposition for only a subset of the training data, we treat the missing decompositions as latent variables. 
We then propose an algorithm based on Hard-EM \cite{Neal98aview}  for learning the model. The rest of this section provides more details.  

\paragraph{Probabilistic Model.}
\label{sec3-prob-model}
Given a question $Q$ and a $C$ context pair, our system generates the answer $A$ according to the following probabilistic model: 

% \vspace{-5mm}
{
\begin{eqnarray}
&&P_{\theta}(A|Q,C) = \sum_Z P_{\theta}(A,Z|Q,C)\\ \label{eq:whole}
 && = \sum_Z P^{\text{dc}}_{\text{LM}}(Z|Q)\times P^{\text{rc}}_{\text{LM}}(A|Q,C,Z) \label{eq:split}
\end{eqnarray}
}

% \vspace{-4mm}
\noindent where $Z$ denotes the unobserved decomposition of the question, $P^{\text{dc}}_{\text{LM}}(Z|Q)$~\footnote{We have made the following independence assumption: $P^{\text{dc}}_{\text{LM}}(Z|Q) \approx P^{\text{dc}}_{\text{LM}}(Z|Q,C)$.} denotes the question decomposer (operationalised based on one specific LM), and $P^{\text{rc}}_{\text{LM}}(A|Q,C,Z)$ denotes the reading comprehension component.
In principle, the $P^{\text{dc}}_{\text{LM}}$ and $P^{\text{rc}}_{\text{LM}}$ components can be constructed using different models, so the parameters $\theta$ of the whole probabilistic model consists of those for these two models. This is denoted by the \emph{separate} variant. 

We further investigate using the same LM for both the question decomposer and reading comprehension component, which we denote by the \emph{unified} variant in the experiments. In this case, the probabilistic model parameter $\theta$ consists of only one set of parameters corresponding to the underlying model.

\paragraph{Question Decomposer.}
To obtain high-quality sub-questions, we first train a question decomposer $P^{\text{dc}}_{\text{LM}}$ to break down difficult multi-hop questions, i.e., the first term in Equation \ref{eq:split}.
It learns the decomposition based on {QDMRs} \cite{break-it-down}.  
We only use the specific partition on the DROP dataset~\cite{DROP:2019} and treat QDMRs as sub-questions.
These sub-questions only cover around 10\% QA pairs in DROP.
Therefore, we need to predict decompositions for the rest of the dataset.
More details will be revealed in Section \ref{sec4-exper}.

Formally, given a multi-hop question $Q$, the question decomposer $P^{\text{dc}}_{\text{LM}}$ generates the sub-questions $Z := \{Q^1, Q^2, ..., Q^s\}$. 
Intuitively, We treat it as a seq2seq learning problem:
our input to the encoder is ``$\texttt{<PARSE>} Q $'', where \texttt{<PARSE>} is a special token. 
The decoder then generates tokens of the sub-questions in auto-regressive way  ``$\texttt{<subQ>} Q^1 \texttt{<subQ>} Q^2 \texttt{<subQ>} \ldots Q^s$'', where \texttt{<subQ>} is a special token~\footnote{We employ the greedy search algorithm to generate the sub-questions $Z$. However, one can leverage other strategies like beam search to make more than one predictions.}. 

\paragraph{Reading Comprehension Component.}
To further obtain answers based on the question and generated sub-questions, the reading comprehension component $P^{\text{rc}}_{\text{LM}}$ generates the answer $A$, i.e., the second term in Equation \ref{eq:split}. 
In stead of directly answering all the sub-questions given by the trained question decomposer, we train our RC component to predict the final answer in a sequence-to-sequence way.

Formally, given a multi-hop complex question $Q$ and the corresponding sub-questions $Z := \{Q1, Q2, ..., Q^s\}$ generated by a trained question decomposer, our input to the RC encoder is ``{\small $\texttt{<QUESTION>} Q \texttt{<subQ>} Q^1 \ldots \texttt{<subQ>} Q^s \texttt{<CONTEXT>} C$}'', where \texttt{<QUESTION>} and \texttt{<CONTEXT>} are special tokens.
In other words, we concatenate the multi-hop question and all the sub-questions, together with the context as the input to our RC component.
The decoder then generates the tokens of the answer autoregressively. 

\begin{algorithm}[t]
\caption{Learning with Hard-EM }\label{alg:cap}
\begin{algorithmic}[1]
\Require an initial pre-trained LM $M$; the full reading comprehesion dataset $\mathcal{D}_1$; the subset with sub-question annotations $\mathcal{D}_2$. 
\State Train $M$ on $\mathcal{D}_2$ to get $M^0$
\For{ iter \textbf{in} N\_iters}
    \State For all $\mathcal{D}=\mathcal{D}_1\setminus\mathcal{D}_2$ employ $M^{iter-1}$ to predict sub-questions and get $\mathcal{D}^{iter}$
    \State Retrain $M^{iter-1}$ on all examples: $\mathcal{D}_2\cup\mathcal{D}^{iter}$, get updated model $M^{iter}$
\EndFor
\end{algorithmic}
\end{algorithm}

\paragraph{Training and Inference.} 
The training objective of our model is

\begin{equation}
\begin{split}
    \mathcal{L} = &\sum_{(Q,C,A) \in \mathcal{D}_1 \setminus \mathcal{D}_2} \log P_{\theta}(A|Q,C) + \\ &\sum_{(Q,C,Z^*,A)\in \mathcal{D}_2} \log P_{\theta}(A,Z^*|Q,C),
\end{split}
\end{equation}

\noindent where $Z^*$ denotes the ground truth decomposition available only for the subset of the training data referred to by $\mathcal{D}_2$.
The first term of the training objective involves  enumerating over all possible latent decompositions, which is computationally intractable. Therefore, we resort to Hard-EM for learning the parameters of our model (see Algorithm  \ref{alg:cap}) for the unified variant. 
We found taking 10 iterations of the Hard-EM algorithm to be mostly sufficient for learning model parameters in our experiments. 

For the separate variant, i.e., using two different LMs for $P^{\text{dc}}_{\text{LM}}$ and $P^{\text{rc}}_{\text{LM}}$, we train the question decomposer on $\mathcal{D}_2$, and then train the reading comprehension component on $\mathcal{D}_2$ as well as $\mathcal{D}_1 \setminus \mathcal{D}_2$ augmented with the generated decomposition $Z$. 
We also compare with training the reading comprehension component on $\mathcal{D}_2$ only, in the experiments. 
During inference time, we first generate the question decomposition $\tilde{Z}$ according to 
$P^{\text{dc}}_{\text{LM}}$, and then use $\tilde{Z}$ in $P^{\text{rc}}_{\text{LM}}$ to generate the answer.

%% file: sec4-exper.tex
%!TEX ROOT = ./acl2021.tex
\section{Experiments}
\label{sec4-exper}

\begin{table}[t]%{0.49\textwidth}
    \centering
    \resizebox{.48\textwidth}{!}{
    \begin{tabular}{c|c||c|c|c|c|c} \hline
       \multicolumn{2}{c||}{Proportions} & 1\% & 5\% & 10\% & 50\% & 100\% \\ \hline \hline
       \multicolumn{2}{c||}{BLEU} & 39.08 & 44.76 & 47.74 & 50.12 & \textbf{54.69}  \\
       \multicolumn{2}{c||}{Rouge-1} & 77.49 & 81.75 & 83.12 & 84.76 & \textbf{85.67} \\
       \multicolumn{2}{c||}{Rouge-2} & 57.00 & 62.83 & 64.97 & 66.94 & \textbf{68.61} \\
       \multicolumn{2}{c||}{RougeL} & 67.78 & 72.65 & 74.37 & 76.55 & \textbf{77.43} \\ \hline
       \multirow{2}{*}{RC} & EM & 26.0 & 26.5 & 27.0 & \textbf{27.8} & 27.2 \\ 
       & F1 & 31.3 & 31.3 & 31.6 & \textbf{32.2} & 32.0 \\ \hline
    \end{tabular}}
    \caption{Experimental results of the Bart based question decomposer: (1) Row 1-4 show intrinsic metrics for the question decomposition by using different proportions of training instances. (2) Row 5-6 show extrinsic metrics of the RC model by using the corresponding decomposer generated sub-questions.}
    \label{tab:bart decomposer}
\end{table}

% \hfill
\begin{table*}[t]
    \centering
    \begin{tabular}{c|c||c|c|c|c|c|c|c|c|c|c} \hline
       \multicolumn{2}{c||}{LMs} & \multicolumn{5}{c|}{t5-small} & \multicolumn{5}{c}{t5-base} \\ \hline
       \multicolumn{2}{c||}{Proportions} & 1\%~\footnote{Trained as a question decomposer, the t5-small model and cannot be further evaluated on downstream RC task, as the generated sub-questions are poor-quality.} & 5\% & 10\% & 50\% & 100\% & 1\% & 5\% & 10\% & 50\% & 100\% \\ \hline \hline
       \multicolumn{2}{c||}{BLEU} & 11.21 & 44.50 & 50.44 & 60.15 & \underline{62.73} & 34.86 & 52.98 & 57.3 & 62.18 & \textbf{64.40}  \\
       \multicolumn{2}{c||}{Rouge-1} & 43.00 & 76.93 & 81.53 & 87.25 & \underline{88.59} & 70.66 & 84.16 & 85.77 & 88.50 & \textbf{89.27} \\
       \multicolumn{2}{c||}{Rouge-2} & 28.18 & 59.13 & 64.33 & 72.60 & \underline{74.76} & 50.57 & 66.86 & 70.24 & 74.24 & \textbf{75.72} \\
       \multicolumn{2}{c||}{RougeL} & 39.22 & 68.92 & 73.66 & 79.99 & \underline{81.57} & 62.10 & 75.49 & 78.07 & 81.20 & \textbf{82.53} \\ \hline
       \multirow{2}{*}{RC} & EM & - & 28.9 & \underline{29.9} & 29.0 & 29.0 & 33.7 & 34.3 & 34.3 & 34.6 & \textbf{34.8} \\ 
       & F1 & - & 33.0 & \underline{34.0} & 33.2 & 33.1 & 37.8 & 38.4 & 38.5 & 38.5 & \textbf{38.6} \\ \hline
    \end{tabular}
    \caption{Results of the T5 based question decomposer (left-half: t5-small, right-half: t5-base): (1) Row 1-4 show all intrinsic metrics to evaluate the question decomposer by using different proportions of training instances. (2) Row 5-6 show extrinsic metrics of the RC component by using the corresponding decomposer generated sub-questions.}
    \label{tab:t5 decomposer}
\end{table*}

\subsection{Dataset}
We consistently use the same notations as in Algorithm \ref{alg:cap}.
\begin{itemize}
    \item $\mathcal{D}_1$: the \textsc{DROP} dataset~\cite{DROP:2019} that contains 77,400/9,536 question ($Q$) answer ($A$) training/testing pairs for the reading comprehension component.
    \item $\mathcal{D}_2$: the \textsc{Break} dataset~\cite{break-it-down}~\footnote{The full \textsc{Break} dataset \citet{break-it-down} annotated is a combination of many datasets including \textsc{DROP}. In this paper, we only use the DROP partition of the original \textsc{Break}.} that contains 7,683/1,268 question ($Q$) decomposition ($Z^*$) training/testing pairs for the question decomposer~\footnote{This subset of DROP contains the corresponding answers for each question. Therefore, we also use it to evaluate the RC component in our experiments.}.
    \item $\mathcal{D} = \mathcal{D}_1 \setminus \mathcal{D}_2$: the difference set between $\mathcal{D}_1$ and $\mathcal{D}_2$ that contains only question answer pairs without ground-truth decomposition.
    \item $\mathcal{D}^{iter}$: $\mathcal{D}$ with decomposition ($Z$) generated by the trained question decomposer.
\end{itemize}
Note that every question ($Q$) is associated with a specific context ($C$). 
With all question decomposition labelled, $\mathcal{D}_2$ is actually a subset of $\mathcal{D}_1$ and is more challenging.

\subsection{Backbone and Evaluation Metric}
There are three LMs of different types and sizes we employ as backbones in this paper: (1) t5-small (60M parameters), (2) t5-base (220M parameters), (3) bart-base (140M parameters).
We also employ GPT-3 (175B parameters) as it is the current state-of-the-art language model in a various of natural language processing tasks.

\noindent\textbf{Sub-question Decomposition}
We train and evaluate our question decomposer using $\mathcal{D}_2$, which was proposed to better understand difficult multi-hop questions.
We report BLEU~\cite{Papineni02bleu:a} and Rouge~\cite{lin-2004-rouge} scores to show the intrinsic performance of the decomposer.

\noindent\textbf{Reading Comprehension}
We evaluate our RC model on $\mathcal{D}_2$.
For the Hard-EM approach, we have $\mathcal{D}_1 \setminus \mathcal{D}_2$ as weakly supervised data.
We report F1 and Exact Match(EM)~\cite{DROP:2019} scores in the following experiments.

\begin{table*}[t]%{0.5\textwidth}
    \centering
    \begin{tabular}{l|c|c||c|c} \hline
       Backbone & Variant & Training Set & F1 & EM\\ \hline \hline
       baselines & & & &\\
       bart-base~\cite{lms-bart} & - & $\mathcal{D}_2$ & 30.9 & 27.1 \\
       t5-base~\cite{lms-t5} & - & $\mathcal{D}_2$ & 37.9 & 33.9 \\ \hline \hline
       our bart-base variants & & & &\\
       w/ predicted sub-questions & \emph{separate} & $\mathcal{D}_2$ & 32.0 & 27.2 \\
       w/ ground-truth sub-questions & \emph{separate} & $\mathcal{D}_2$ & 33.2 & 29.0 \\
       w/ ground-truth sub-questions & \emph{separate} & $\mathcal{D}_2, \mathcal{D}^{1}$ & \textbf{45.0} & \textbf{40.5} \\
       w/o Hard-EM & \emph{unified} & $\mathcal{D}_2, \mathcal{D}^{1}$ & 44.2 & 39.9 \\
       w/ Hard-EM & \emph{unified} & $\mathcal{D}_2, \mathcal{D}^{iter}$ & \underline{44.3} & \underline{40.0} \\ \hline \hline
       our t5-base variants & & & &\\
       w/ predicted sub-questions & \emph{separate} & $\mathcal{D}_2$ & 38.6 & 34.8 \\
       w/ ground-truth sub-questions & \emph{separate} & $\mathcal{D}_2$ & 39.6 & 35.6 \\
       w/ ground-truth sub-questions & \emph{separate} & $\mathcal{D}_2, \mathcal{D}^{1}$ & \textbf{45.1} & \textbf{40.8} \\ %49.8, 45.9
       w/o Hard-EM & \emph{unified} & $\mathcal{D}_2, \mathcal{D}^{1}$ & 38.8 & 34.9 \\
       w/ Hard-EM & \emph{unified} & $\mathcal{D}_2, \mathcal{D}^{iter}$ & \underline{44.0} & \underline{40.1} \\ \hline \hline
       GPT-3 (zero-shot) & - & - & 15.7 & 4.6 \\ 
       GPT-3 (few-shot) & - & - & 34.9 & 27.0 \\ \hline
    \end{tabular}
    \caption{Overall results for baselines, our separate and unified variants. All models are evaluated on the same test set from $\mathcal{D}_2$.}
    \label{tab:models}
\end{table*}
% \caption{Experimental results.}
% \vspace{-8pt}
% \end{table*}

\subsection{Results on Decomposition}
Based on Bart and T5, Table \ref{tab:bart decomposer} and Table \ref{tab:t5 decomposer} respectively show the experimental results of the question decomposers.
To comprehensively show their performance, we conducted two aspects of experiments including intrinsic decomposition evaluation and extrinsic RC evaluation.

\paragraph{Intrinsic Evaluation}
We first evaluate the quality of sub-questions generated by different question decomposers. 
In this part, intrinsic metrics, BLEU and Rouge scores, are shown in the first four rows of Table \ref{tab:bart decomposer} and Table \ref{tab:t5 decomposer}.
And also we show the results of five decomposers trained on different proportions (1\%, 5\%, 10\%, 50\%, 100\%) of the \textsc{Break} dataset $\mathcal{D}_2$'s training data.
All these evaluations are conducted on the same validation set of $\mathcal{D}_2$.

Comparing column-by-column, we find that with more training data, both question decomposers achieve a better performance for both BLEU and Rouge.
We also note that the rate of improvement of these metrics becomes slower when more data is added (e.g.\ 1\% to 5\% and 10\% to 50\%).
Therefore, we posit that with more training data, the performance of the decomposer will not improve due to the capability of the LM model.

\paragraph{Extrinsic Evaluation}
Since the eventual usage of the generated sub-questions is to improve the RC component, we conduct a RC performance comparison experiments to see how can the quality of these sub-questions influence the downstream RC task.
Also like the intrinsic evaluation, we show the results based on decomposers trained on different proportions of $\mathcal{D}_2$ by using two extrinsic metrics: EM and F1.
All the evaluations are conducted on the same validation set of $\mathcal{D}_2$.

To clarify our settings in this part, we don't employ the ground-truth sub-questions from $\mathcal{D}_2$.
Instead, we employ the sub-questions generated by five question decomposers for the RC component to predict answers.
As the last two rows of both Table \ref{tab:bart decomposer} and Table \ref{tab:t5 decomposer} show, both EM and F1 scores show a gradually increasing trend when more training instances are used to train the question decomposer.
With more parameters, t5-base tends to have a better performance than t5-small.

\subsection{Results on Reading Comprehension}
Table \ref{tab:models} shows the experimental results for the downstream RC task.
We show two baselines in the first place: ``bart-base'' and ``t5-base''.
Without taking sub-questions as input, both are trained on the \textsc{Break} dataset $\mathcal{D}_2$.
Based on these vanilla models, we show our \emph{separate} and \emph{unified} approaches that use ``bart-base'' and ``t5-base'' as backbones separately in Table \ref{tab:models}.

\subsubsection{Separate Variant}
Our \emph{separate} variants are base on the architecture in Figure~\ref{fig:decomposer-rc}.
In Table \ref{tab:models}, we have three \emph{separate} variants based on each backbone for comparison.
Taking t5-base as one example, comparing to the t5-base, using predicted sub-questions achieves a 0.7-point gain of F1 score.
Meanwhile using ground-truth sub-questions, our model outperforms the t5-base by 1.7 points of F1 score.
The same improvement can be also concluded from the bart-base model.
They employ $\mathcal{D}_2$ for training but their testing sets are different: predicted one use generated sub-questions while ground-truth one use sub-questions from $\emph{D}_2$.
The reason why our approach is more effective than the baseline model is that concatenating sub-questions can give LMs hints on the reasoning procedure, which helps LMs produce step-by-step thoughts implicitly.

Furthermore, we add $\mathcal{D}^1$ as the training set to train our seperate model.
As it shows in Table \ref{tab:models}, this kind of \emph{separate} variants show the overall best performance since we have two sets of parameters separately learning question decomposition and reading comprehension.
Compared to t5-base, the bart-base variant shows a higher performance gain that proves the effectiveness of our method.

\begin{table*}[t]
    \centering
    \begin{tabular}{p{0.41\linewidth}|p{0.18\linewidth}|p{0.1\linewidth}|p{0.1\linewidth}|p{0.09\linewidth}}
    \hline
        Context & Question & GPT-3 (few-shot) & bart-base \emph{separate} (best) & ground-truth answer\\ \hline 
        ... notably striking out \textcolor{cyan}{Julio Franco}, at the time the \textcolor{cyan}{oldest player} in the MLB at \textcolor{cyan}{47 years old}; \textcolor{cyan}{Clemens} was himself \textcolor{cyan}{43}. In the bottom of the eighteenth inning, Clemens came to bat again... & Which player playing in the 2005 National League Division Series was older, Julio Franco or Roger Clemens? & Julio Franco (\cmark) & Julio Franco (\cmark) & Julio Franco \\ 
        \hline 
        ... Nyaungyan then systematically reacquired nearer Shan states. He \textcolor{cyan}{captured Nyaungshwe in February 1601, and the large strategic Shan state of Mone in July 1603}, bringing his realm to the border of Siamese Lan Na. In response, Naresuan of Siam marched in early 1605 to ... & How many years after capturing Nyaungshwe did Nyaungyan capture the large strategic Shan state of Mone? & 3 years (\xmark) & 2 (\cmark) & 2 \\
        \hline 
        Kannada language is the official language of Karnataka and spoken as a native language by about \textcolor{red}{66.54\%} of the people as of 2011. Other linguistic minorities in the state were Urdu (10.83\%), \textcolor{cyan}{Telugu language (5.84\%)}, Tamil language (3.45\%), ... & How many in percent of people for Karnataka don't speak Telugu? & 66.54\% (\xmark) & 94.04\% (\xmark) & 94.16\% \\
        \hline 
        A 2013 analysis of the National Assessment of Educational Progress found that \textcolor{cyan}{from 1971 to 2008, the size of the black-white IQ gap in the United States decreased from 16.33 to 9.94 IQ points}. It has also concluded however that, ... & How many IQ points did the black-white IQ gap decrease between 1971 and 2008? & 16.33 (\xmark) & 0.9 (\xmark) & 6.39\\ 
        \hline
    \end{tabular}
    \caption{Correct and incorrect outputs from GPT-3 and our \emph{separate} variant. \textcolor{cyan}{Correct} and \textcolor{red}{Wrong} supporting facts are annotated in the context using the corresponding color. Correct and wrong answer predictions are also marked with \cmark and \xmark \ (the table is best seen in colours).}
    \label{tab:error}
\end{table*}

\begin{table*}[t]
    \centering
    \begin{tabular}{l|l||c|c|c|c}
    \hline
        \multicolumn{2}{c||}{overlaps} & $0\sim25\%$ & $25\%\sim50\%$ & $50\%\sim75\%$ & $75\%\sim100\%$ \\ \hline
        \multirow{4}{*}{uni-grams} & bart-base & - & 0 & 25.7 & 27.4 \\
        & \emph{unified} & - & 0 & 32.9 & \underline{40.2} \\
        & \emph{separate} & - & 0 & \textbf{35.7} & \textbf{41.3} \\
        & GPT-3 & - & \textbf{100.0} & \textbf{35.7} & 26.4 \\
        \hline \hline
        \multirow{4}{*}{bi-grams} & bart-base & - & 16.7 & 23.6 & 28.2 \\
        & \emph{unified} & - & 33.3 & \textbf{29.1} & \underline{41.9} \\
        & \emph{separate} & - & \textbf{50.0} & 28.6 & \textbf{43.2} \\
        & GPT-3 & - & \underline{44.4} & \textbf{29.1} & 26.2 \\
        \hline \hline
        \multirow{4}{*}{tri-grams} & bart-base & 22.2 & 20.5 & 25.5 & 29.3 \\
        & \emph{unified} & 38.9 & 26.2 & \underline{32.3} & \underline{45.1} \\
        & \emph{separate} & \textbf{50.0} & \textbf{30.0} & \textbf{33.4} & \textbf{45.9} \\
        & GPT-3 & \textbf{50.0} & \underline{28.0} & 25.8 & 26.8 \\
        \hline
        
        \hline
    \end{tabular}
    \caption{EM scores separately computed based on overlaps of sub-questions n-grams between training set and testing set on $\mathcal{D}_2$. Four models listed in this table are: the bart-base baseline, the best performed \emph{separate} model, the best performed \emph{unified} model}
    \label{tab:n-grams}
\end{table*}

\subsubsection{Unified Variant}
Our \emph{unified} variants are base on the architecture in Figure~\ref{fig:decomposer-rc} and one single model is used to train on both steps.
In Table \ref{tab:models}, the last two rows of each variants show the performance of our \emph{unified} variant.
Without the Hard-EM algorithm, performing multi-task learning achieves a 0.9 point improve over the T5 baseline. However, it shows a performance drop when compared to the \emph{separate} variant with ground-truth sub-questions.
This can be caused by the enlarged dataset and the additional decomposition work the \emph{unified} variant need to handle.

When more training data is provided (i.e.\ $\mathcal{D}^1$ and $\mathcal{D}^{iter}$), though without ground-truth sub-questions, the \emph{unified} variants substantially outperforms the baselines by 10.1 and 6.1 points over bart-base and t5-base model. 
Furthermore, when compared with the best \emph{separate} variants, our \emph{unified} models also show comparable performance on both F1 and EM metrics.
Based on the observations of the last three rows of each backbone, it can be concluded that introducing more weakly-supervised training data can significantly help our model address the original difficult multi-hop RC task.

We also include another evaluation of employing GPT-3, which is the state-of-the-art language model on many tasks and also in a large parameter scale (175B).
The results are shown by last two rows in Table \ref{tab:models}.
Based on the experimental results, GPT-3 cannot even beat two baseline models under the zero-shot learning paradigm, which again shows the complexity and challenging of the task.
When provided with several exemplars, it can easily outperform the bart-base model by 2.4 points on F1 score.
However, even with $\times 1000$ parameters, GPT-3 is still far behind to our best variants by 10.2 F1 points.

\section{Analysis and Discussions}

\subsection{Qualitative Analysis}
In this section, we will further discuss some real-life cases generated by our proposed variants from the dataset.
In Table \ref{tab:error}, the first row shows a comparison question and both GPT-3 and our bart-base \emph{separate} model can produce the correct answer.
However, when the question requires some arithmetic operations, such as addition or subtraction, the GPT-3 model would fail to answer correctly.
Our model can handle this as shown by the second row.

There are two types of failures from our variants: one is that our model cannot handle unseen numbers, and the other is arithmetic between float numbers.
The unseen number case happens in the third row of Table \ref{tab:error}. 
Asking for the number of a complement set, though the number 94.04\% is wrongly predicted by our model, it is more close to the ground-truth (94.16\%) when compared to the GPT-3, which directly predict an wrong evidence annotated with red color.
Furthermore, the last row shows a subtraction question between two float numbers. 
Different from integer number subtraction in the second row, it is much harder to compute this arithmetic for language models.
Traditionally, some symbolic methods can handle this problem very well.
Tackling these problems can be interesting  future work directions.

\subsection{Quantitative Analysis}
We look into details of $\mathcal{D}_2$ from the perspective of sub-question n-grams for both training and testing data.
Intuitively, given one instance from the test set, more n-grams overlap it shows with the training set, higher the EM and F1 scores.
Therefore, we further conducted the analysis and list all the statistics in Table \ref{tab:n-grams}.

We calculate for uni-grams, bi-grams and tri-grams for four models: bart-base baseline, the best-performed \emph{separate} and \emph{unified} variants proposed in Section \ref{sec3-model} and GPT-3 with few-shot learning.
The overlaps we choose is four intervals using percentages to represent.
For example, $0\sim25\%$ overlapping on bi-grams means that the test instance have this proportion of bi-grams overlaps with all the training instances.
Note that there is no overlapping  for uni-grams and bi-grams in $0\sim25\%$.

In Table \ref{tab:n-grams}, we report the EM score (F1 score shows the similar results).
The bart-base model show a tendency that with more overlaps across all n-grams, the performance will increase, which is consistent with our assumption.
However, on the contrary, GPT-3 model show a reverse tendency that is probably due to the pre-trained corpus that shares far less n-grams with the test set.
This characteristic improves the compositional generalisation ability as it outperforms the baseline model on the low-overlapping part of test set.
Both of our \emph{separate} and \emph{unified} variants show overall improvements over the bart-base baseline.
In particular, the first and second columns also show our model can better handle the low-overlapping questions, even without performance drop on the high-overlapping questions ($50\%\sim100\%$).
This experiment can further prove the compositional generalisation of our method is comparable to GPT-3.

%% file: sec5-concl.tex
%!TEX ROOT = ./acl2021.tex
\section{Conclusion}
\label{sec5-concl}
We propose a two-step process for multi-hop reading comprehension task.
The first step involves a question decomposer that maps a difficult multi-hop question into several sub-questions.
The second step is to train a reading comprehension model based on (question, sub-questions, paragraph, answer) tuples.
With the addition of sub-questions, our bart-/t5-base variants outperform the baseline model by 2.3/1.7 using ground truth sub-questions and 1.1/0.7 using generated ones on F1 score.
Based on the hard-EM paradigm, large positive gains of another 11.1/4.4 point on F1 by the unified multi-task learning bart-/t5-base models shows the effectiveness of introducing weakly supervised training data.
By further analysing the predicted examples and dataset, we also found our model can make a more comprehensive improvement compared with the SOTA GPT-3 model.
But some problems like handling unseen numbers still exist and will be our future research directions.